\pdfoutput=1

\documentclass[11pt]{article}

\usepackage[]{EMNLP2023}

\usepackage{times}
\usepackage{latexsym}

\usepackage[T1]{fontenc}

\usepackage[utf8]{inputenc}

\usepackage{microtype}

\usepackage{inconsolata}
\usepackage{graphicx}
\usepackage{booktabs}
\usepackage{multirow}

\title{SAIL: Search-Augmented Instruction Learning}

\author{Hongyin Luo$^1\;$ Yung-Sung Chuang$^1\;$ Yuan Gong$^1\;$ Tianhua Zhang$^2$\\ {\bf Yoon Kim$^1\;$ Xixin Wu$^2\;$ Danny Fox$^3\;$ Helen Meng$^2\;$ James Glass$^1$} \\ $^1$ MIT Computer Science and Artificial Intelligence Lab, Cambridge MA, USA\\ $^2$ CUHK Centre for Perceptual and Interactive Intelligence, Hong Kong SAR, China \\ $^3$ MIT Linguistics, Cambridge MA, USA \\ \url{hyluo@mit.edu} $\;\;$ \url{https://openlsr.org/sail-7b}}

\begin{document}
\maketitle

\begin{abstract}

Large language models (LLMs) have been significantly improved by instruction fine-tuning, but still lack transparency and the ability to utilize up-to-date knowledge and information. In this work, we propose search-augmented instruction learning (SAIL), which grounds the language generation and instruction following abilities on complex search results generated by in-house and external search engines. With an instruction tuning corpus, we collect search results for each training case from different search APIs and domains, and construct a new search-grounded training set containing \textit{(instruction, grounding information, response)} triplets. We then fine-tune the LLaMA-7B model on the constructed training set. Since the collected results contain unrelated and disputing languages, the model needs to learn to ground on trustworthy search results, filter out distracting passages, and generate the target response. The search result-denoising process entails explicit trustworthy information selection and multi-hop reasoning, since the retrieved passages might be informative but not contain the instruction-following answer. Experiments show that the fine-tuned SAIL-7B model has a strong instruction-following ability, and it performs significantly better on transparency-sensitive tasks, including open-ended question answering and fact checking.

\end{abstract}

\section{Introduction}
Large language models (LLMs) have demonstrated many impressive capabilities, including zero-shot inference and few-shot in-context learning \cite{wei2022emergent}.  Recent research has shown that LLMs benefit from instruction tuning \cite{ouyang2022training}, and that such instruction-tuned LLMs significantly outperform plain LLMs on zero-shot language tasks \cite{peng2023instruction}.
Instruction-tuned LLMs have shown an ability to generate both natural and programming languages following natural language guidance and requests. To achieve the same goal, a pretrained LLM needs a number of annotated examples as in-context learning prompts.

Despite their impressive behavior, LLMs have a number of issues, including obsolence and transparency.
Understandably, LLMs are trained with corpora constructed up to a certain time point. With this fixed, pretrained or fine-tuned model, subsequently occurring information cannot appear in any informed generation by the LLM. One way to update the knowledge in LLMs is to re-train the entire model with an updated training corpus. However, this would be costly and time-consuming. 

In terms of transparency, the predictions of LLMs are opaque because generations are not grounded on trustworthy sources. It is possible for an LLM to generate undesirable language that looks like human-generated text, including misinformation, stereotypes, and toxic language~\cite{zhang2023interpretable,hartvigsen2022toxigen}. Without providing legitimate sources for LLM generated texts it is difficult to catch and avoid these undesirable LLM behaviors.

\begin{figure*}[h]
\centering
\includegraphics[width=\textwidth]{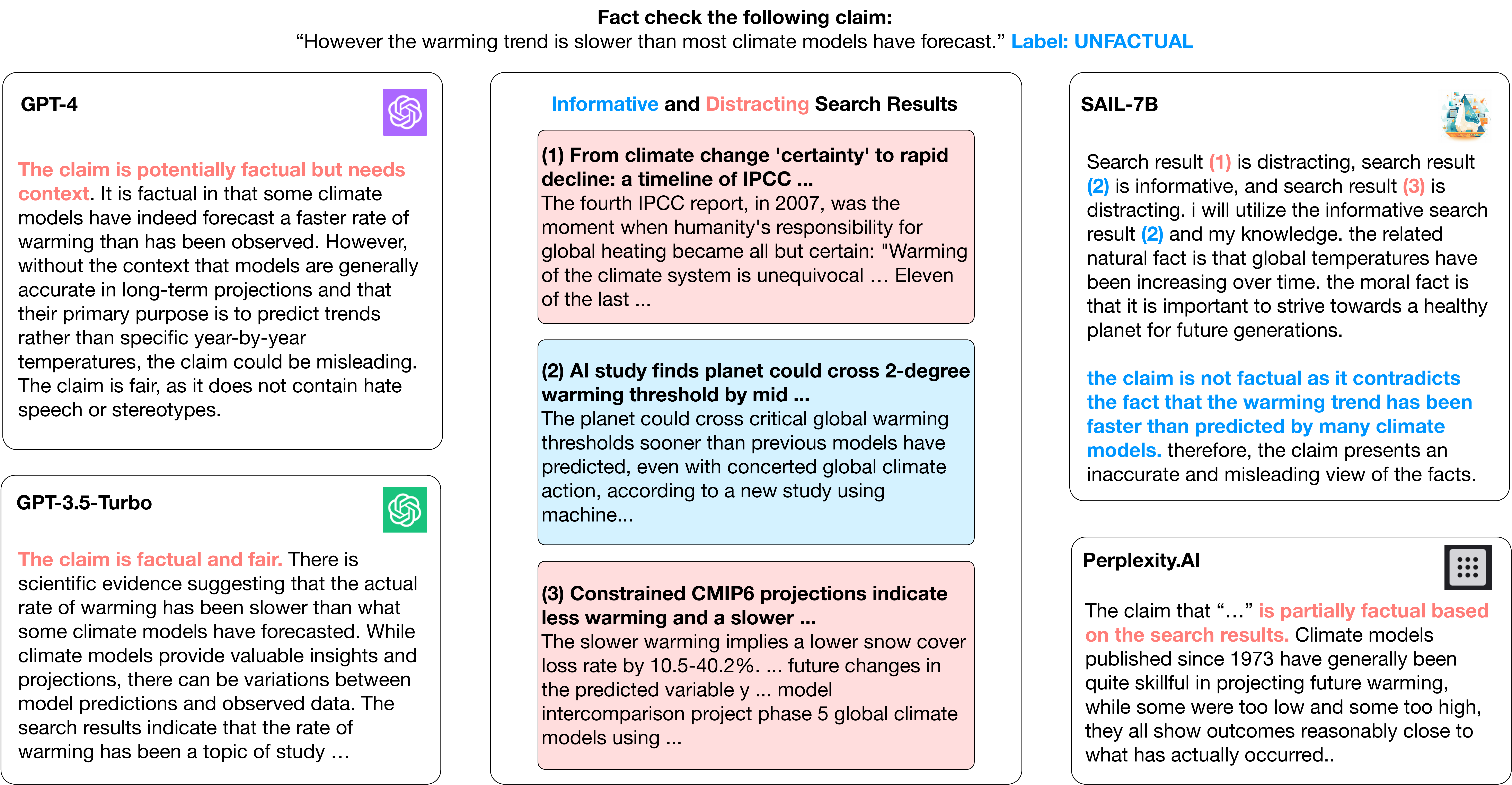}
\caption{Fact checking grounding on complicated search results with \textbf{SAIL-7B} and strong commercial language models. The first and third passages are distracting since they do not contain information that supports or refutes the claim, while the second passage disagrees with the claim. SAIL-7b Successfully make the the correct prediction while other commercial LLMs are distracted.
}
\label{fig:intro}
\end{figure*}

To overcome these difficulties, a straightforward solution is to connect LLMs to information retrieval systems, especially commercial search engines. By doing so, the LLM can ground its predictions on information retrieved from an up-to-date knowledge base, and the sources of the generations would be transparent to users. Before LLMs became large enough to memorize a significant amount of world knowledge, retrieval-based grounding had been heavily studied for open-domain question answering \cite{chen2017reading,kwiatkowski2019natural,guu2020retrieval}. Recent LLMs have also shown the potential of using information retrieval tools, e.g., Toolformer \cite{schick2023toolformer} and the ChatGPT \cite{chatgpt2023} retrieval plugin. However, there remains a challenge: is there a trustworthy retrieval model and knowledge base that can be utilized by LLMs?

Existing studies on open-domain question answering have chosen Wikipedia as the \emph{de facto} knowledge base that contains the answer to most questions. However, \citet{zhang2023interpretable} found that the knowledge contained in Wikipedia is not sufficiently up-to-date nor complete for many tasks that require the latest knowledge, so grounding on Wikipedia might lead to worse answers than fully relying on LLMs. Another option is to leverage an internet search engin such as, for example, Google, Bing, and \href{https://duckduckgo.com/}{DuckDuckGo.com}\footnote{A free, privacy-proof, zero-tracking search engine.}.

Although widely used commercial search engines can index and retrieve a vast range of up-to-date information, their retrieval accuracy is ultimately limited,  and third-party users cannot control the performance at the model level. As a result, retrieval results can be noisy, and unrelated information might be shown to users.  This behavior suggests that there is a trade-off between deploying in-house retrieval systems and external search engines. Although it is possible to prompt LLMs to directly use the retrieval results, distracting search results can mislead the model and negatively influence the model's performance. As shown in Figure \ref{fig:intro}, ChatGPT is confused by a distracting passage and generates an incorrect fact check.

The challenges mentioned above are contradictory, and both have a negative impact on grounded language modeling with current LLMs - static knowledge bases and in-house retrievers are not sufficient or up-to-date for all tasks, while commercial search engines often generate distracting results. To address these challenges simultaneously, we propose a search-augmented instruction learning (SAIL) model. Given input instructions and contexts, the model is trained to generate high-quality responses according to the instruction grounding on the noisy research results. In other words, the model learns to denoise the retrieval results to generate high-quality responses.

In summary, we make the following contributions in this work:
\begin{itemize} \setlength{\itemsep}{0pt} \setlength{\parsep}{0pt}
    \item[1.] We show that instruction-tuned LLMs can be heavily misled by distracting grounding information and noisy search results.
    \item[2.] We constructed a search-augmented instruction training corpus.
    \item[3.] We fine-tune a 7B-parameter language model (\texttt{SAIL-7B}) with the constructed training set, which outperforms strong baseline models including \texttt{GPT-3.5-Turbo} and \texttt{Vicuna-13B} on several NLP tasks.
\end{itemize}
By comparing the \texttt{SAIL-7B} model with \texttt{LLaMA-7B}, \texttt{Vicuna-7B}, \texttt{GPT-3.5-turbo}, and \texttt{Vicuna-13B} models on instruction following, question answering, and language checking tasks, we find that the \texttt{SAIL-7B} model has a strong instruction following ability and is robust against distracting grounding search results generated by different retrieval models. In addition, the SAIL model also achieves comparable performance to state-of-the-art instruction-following LLMs.

\section{Method}

\subsection{Search Result Collection}
In this work, we use the 52k self-instruction corpus created by the Alpaca team \cite{alpaca}, and the corresponding responses generated by GPT-4 \cite{peng2023instruction}. For each instruction, we construct a search query by simply concatenating the instruction and the input, if any, and truncating the query to at most 60 words to fulfill the limitation of the search engine. 

The constructed queries are fed into the DuckDuckGo search engine and the BM25 Wikipedia retriever, and the top three search results are retained. Each result consists of three fields: the title, a short piece of preview text, and the corresponding URL of the webpage. For simplicity, we do not further scrape the retrieved webpage, but just use the title and preview texts for further processing.

Each training example is assigned a different search result. We pool the top-three DuckDuckGO and top-two BM25 search passages, a total of five search results. Among this pool, we randomly sample zero, one, two, and three search results with 20\%, 20\%, 20\%, and 40\% probability. Given this randomness, some training cases could be associated with search results from a single source.

\subsection{In-context Retrieval Selection}
To encourage the LLM to focus on trustworthy and informative search results, we concatenate a search filtering sequence before each annotated response. For example, \textit{``Search result (1) is informative and search result (2) is distracting, so I will use the information from the search result (1).''}

However, the trustworthiness of each search result is not labeled, and the number of retrieval items is large. To solve this problem, we employ an entailment classification model proposed in \cite{luo-glass-2023-logic}. We feed each retrieved passage and the corresponding response into the entailment model and compare the entailed and contradictory scores. While most predictions are neutral against the response, the relation between entailed and contradictory scores can roughly indicate if a retrieved passage can provide useful information to generate the target response. As a result, we label \textit{``search result (i) is informative''} if the entailed score is higher than the contradiction score, otherwise the search item is distracting. With the constructed label responses, the \texttt{SAIL-7b} model can generate in-context search selection sequences as shown in Figure \ref{fig:intro}.

\subsection{Fine-tuning}

After collecting the search results and generating in-context retrieval selection sequences, we construct input prompts following Figure \ref{fig:prompts} (b) with GPT-4 generated responses \cite{peng2023instruction}. Note that the most relevant retrieval result is located at the closest position to the instruction for the model to better use its information. We fine-tune both \texttt{LLaMA-7b} models with the constructed prompts to generate both in-context retrieval selection and annotated responses.

In practice, the models are fine-tuned with academic devices.
Specifically, we use 4 $\times$ NVIDIA RTX A6000 GPUs (48GB $\times$ 4) to train the models for 3 epochs. We apply mixed-precision training (fp16) with the standard AdamW optimizer.  We set the maximum sequence length as 1,600 and the batch size as 32. Following Vicuna, we apply gradient checkpointing to reduce the memory cost. The entire fine-tuning process takes 24 hours (24 $\times$ 4 GPU hours). To enable the fine-tuning, we applied gradient offload with Deepspeed and full-sharded data parallel (FSDP) \cite{paszke2019pytorch}.

\begin{figure}[t]
\centering
\includegraphics[width=\columnwidth]{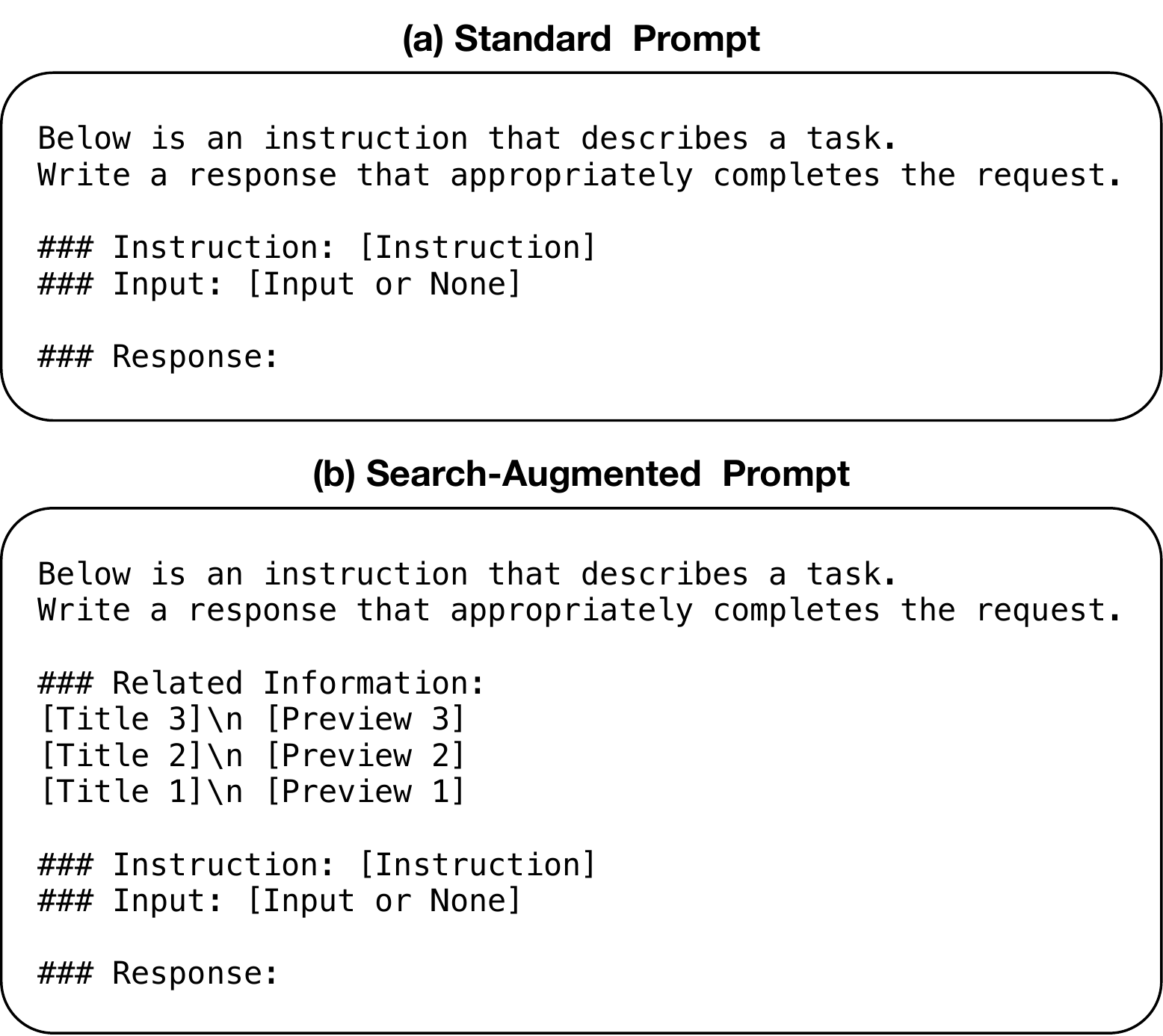}
\caption{Different prompting strategies used in this work. (a) \textbf{Standard prompt}: the prompt template used in \citet{peng2023instruction} to generate GPT-4 responses to the 52k instructions. (b) \textbf{Search-augmented prompt}: combining the top three search results and the instruction.
}
\label{fig:prompts}
\end{figure}

\subsection{Evaluation}
\noindent \textbf{SAIL for instruction following.} Following \citet{peng2023instruction}, we evaluate the instruction following the quality of different models by comparing with GPT-4 responses on the same set of instructions and scoring with GPT-4.

For each case, we construct an evaluation prompt by concatenating the instruction, the GPT-4 response, and the response of the target model. We feed the evaluation prompt to GPT-4 and ask it to score the two responses between 0 to 10. We use the Vicuna-Instructions-80\footnote{\url{https://github.com/lm-sys/FastChat/blob/main/fastchat/eval/table/question.jsonl}} corpus \cite{vicuna2023}, which contains 80 questions to evaluate all models and we calculate the total score a model receives on all questions. We use the evaluation prompt authored by the Vicuna team\footnote{\url{https://github.com/lm-sys/FastChat/blob/main/fastchat/eval/table/prompt.jsonl}}. The highest possible score is $80\times10=800$. It is worth noting that GPT-4 responses can receive slightly different scores against different counterparts. To normalize the difference, we calculate the ratio of model score / GPT-4 score for each test case as the final assessment as implemented in \citet{peng2023instruction}.\\

\begin{figure*}[t]
\centering
\includegraphics[width=\textwidth]{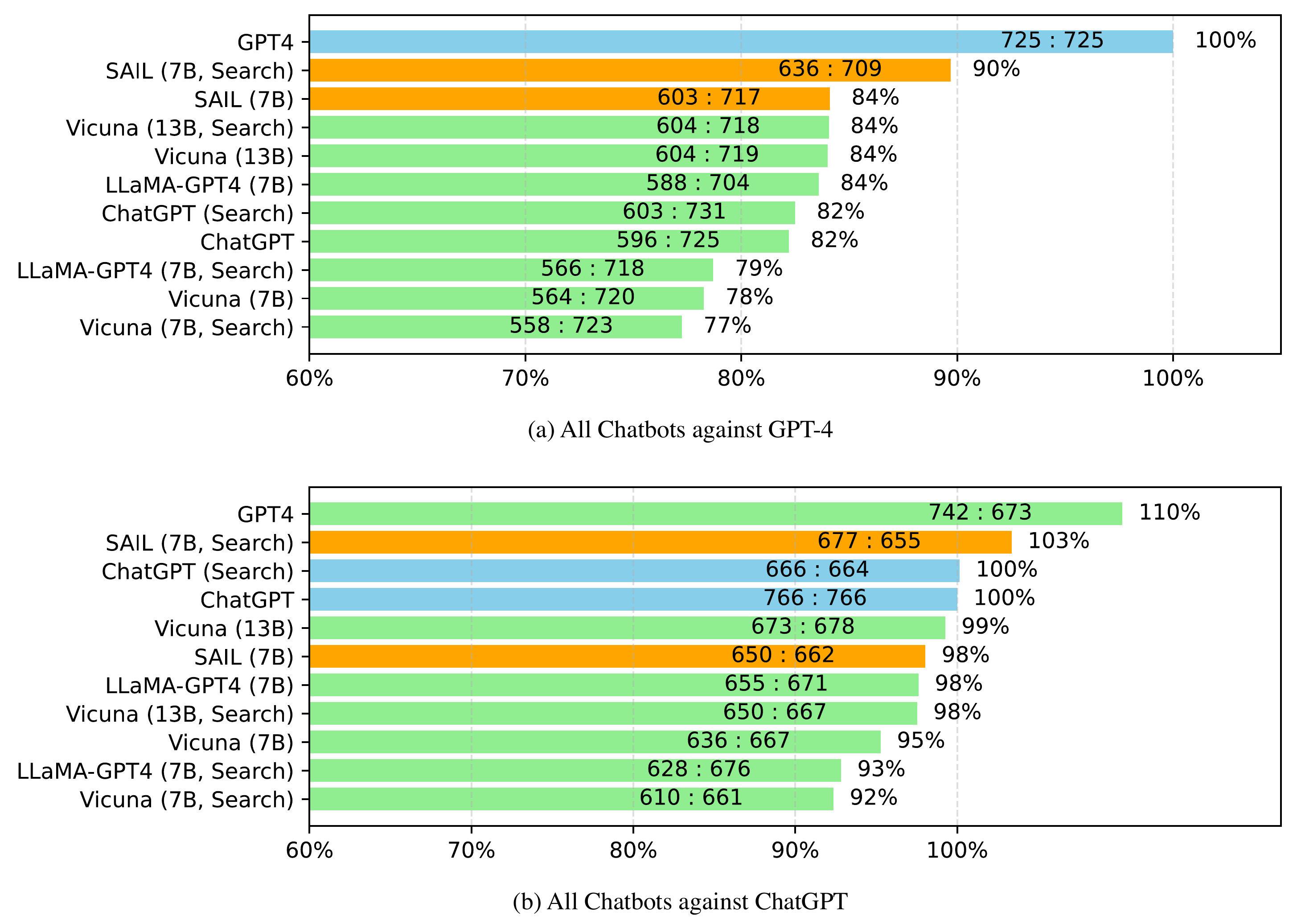}
\caption{Scoring results of all language models on the instruction-following benchmark against GPT-4 and ChatGPT. \textbf{Search} indicating generating responses with language models grounding on search results retrieved by \texttt{DuckDuckGO}, and \textbf{SAIL (7B)} stands for generating responses without search results, although the model is trained for grounded generations. Both Vicuna-7\&13B are version 1.1 models.}
\label{fig:inst_score}
\end{figure*}

\noindent \textbf{SAIL for Question Answering.}
Besides evaluating the quality of instruction-guided generations, we also assess the model's ability to answer commonsense questions. We also test the models on two different settings, including instructed zero-shot prediction and the search-augmentation mode. We evaluate the model performance on CommonsenseQA (CSQA; \citet{talmor-etal-2019-commonsenseqa}), OpenbookQA (OBQA; \citet{Mihaylov2018CanAS}), and ARC-Challenge \cite{Clark2018ThinkYH} benchmarks. Both tasks require answering open-ended questions by selecting from a given set of candidate answers. Through the question-answering experiments, we show that instruction-tuned language models can be significantly biased by noisy research results.\\

\noindent \textbf{SAIL for Fact and Fairness Checking.}
With the recent advances in LLMs that generate human-like languages without guaranteed alignment, human and machine-generated misinformation, stereotypes, and toxicity have become timely and significant concerns. Recent studies have shown that with appropriate instructions and prompts, LLMs can perform unified fact and fairness checking \cite{zhang2023interpretable}. However, other attempts have relied only on LLMs, without grounding on any external sources, thus reducing the trustworthiness and transparency of the checking results.

In this work, we evaluate instructed fact and fairness checking, with the UniLC benchmark proposed in \cite{zhang2023interpretable}, including Climate-Fever, PubHealth, Hate Speech Detection, and Social Biase Frame (SBIC) tasks with two different settings - zero-shot and search-augmented. While we are not aware of what corpora are used to train GPT-4 and ChatGPT, we assess the language-checking performance of \texttt{Vicuna-7B-v1.1}, \texttt{Vicuna-13B-v1.1}, and \texttt{SAIL-7B} with and without search results.

\section{Experiments}
\begin{figure*}[t]
\centering
\includegraphics[width=.95\textwidth]{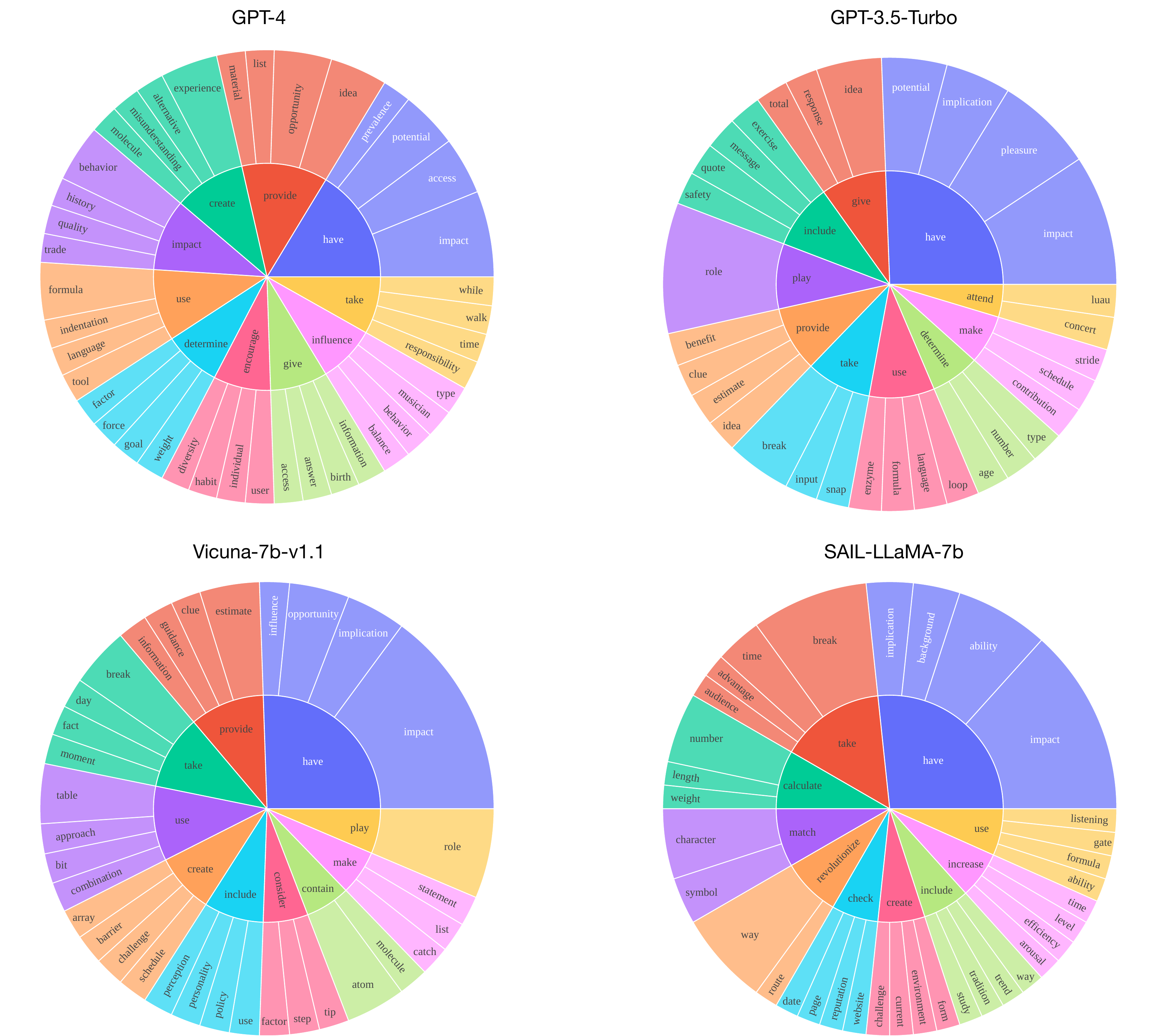}
\caption{Top-10 verbs and associated nouns generated by selective large language models.}
\label{fig:verb-noun}
\end{figure*}
\subsection{Instruction Following}
\noindent\textbf{Automatic Evaluation with GPT-4.} We compare the performance of different models under end-to-end and search grounded settings against GPT-4 and ChatGPT models. The scoring results are shown in Figure \ref{fig:inst_score}.

By comparing to GPT-4, we find that the search-augmented \texttt{SAIL-7B} model significantly outperforms all other models (90\% vs <85\%) using fewer training instructions and parameters, including strong baselines including \texttt{Vicuna-13B} and \texttt{GPT-3.5-turbo} powered ChatGPT. This indicates that when the grounding information is provided, the model does not need as many parameters to memorize knowledge. In addition, the \texttt{SAIL-7B} model also achieves high performance even without search results, showing that the model performance is stable under different generation settings. Similar conclusions can be found by comparing all models against ChatGPT. While GPT-4 is still better, experiment results show that the search-augmented \texttt{SAIL-7B} model achieves 103\% of ChatGPT performance and the no-augmentation SAIL model achieves 98\%, outperforming several strong baselines, including LLaMA tuned on GPT4 instructions and Vicuna models with the same number of parameters. Besides GPT-4, search-augmented \texttt{SAIL-7B} is the only model that outperforms ChatGPT on both experiments.

In addition, we found that the search augmentation makes a significantly higher positive contribution to the SAIL model than all other models. With ChatGPT, the effect of feeding search-augmented prompts with instructions leads to very slight improvements in both evaluations. However, grounding on search results can hurt the performance of Vicuna and LLaMA-GPT4 models of different sizes. By comparing against GPT4, \texttt{Vicuna-13B} is slightly improved by search results, but the improvement is not present when compared to ChatGPT. For the \texttt{Vicuna-7B} and \texttt{LLaMA-7B-GPT4} baselines, augmenting input prompts with search engine outputs makes a significant, negative impact on both evaluations. On the other hand, applying search augmentation to SAIL-7B significantly improves model performance on both experiments (84\% to 90\% and 98\% to 103\%). These results inform our findings:
\begin{itemize} \setlength{\itemsep}{0pt} \setlength{\parsep}{0pt}
    \item The search results contain useful information that can improve the performance of instruction-following language models.
    \item Without search-augmented fine-tuning, it is difficult for a language model to utilize valuable information among the complicated search results, and distracting retrieval results can mislead the generations.
    \item Search-augmented instruction learning can help the model better utilize the valuable information among noisy search results and improve instruction-following performance.
\end{itemize}

\noindent\textbf{Data Statistics.} We first show the word preference of different models on the 80 unseen instructions. The results are shown in Figure \ref{fig:verb-noun}. We compare the distributions of top-10 verbs generated by \texttt{GPT4}, \texttt{GPT-3.5-Turbo} (ChatGPT), \texttt{Vicuna-7B-v1.1}, and \texttt{SAIL-7B} models. With search augmentation, \texttt{SAIL-7B} generates significantly more verbs that do not overlap with GPT's generations, as shown in Table \ref{tab:spe-verbs}. Only two top-10 verbs generated by Vicuna are not covered by GPT-4 and ChatGPT, while six out of ten verbs generated by SAIL-7b are not high-frequency verbs by the GPT models. This indicates that the grounding search results can shift the generation preference of the language models.
\begin{table}[]
\centering
\small
\begin{tabular}{@{}lll@{}}
\toprule
\textbf{Models} & Vicuna-7B-v1.1 & SAIL-7B       \\ \midrule
\textbf{Novel}  & Include        & Calculate     \\
\textbf{Verbs}  & Consider       & Match         \\
                &                & Revolutionize \\
                &                & Check         \\
                &                & Include       \\
                &                & Increase      \\ \midrule
\textbf{Count}           & 2              & \textbf{6}    \\ \bottomrule
\end{tabular}
\caption{Top-10 verbs generated by LLaMA-based models that do not overlap with GPT-4 and ChatGPT.
}
\label{tab:spe-verbs}
\end{table}

The statistics of the generated responses is shown in Table \ref{tab:length}. GPT-4 generates the longest and most diverse responses, while ChatGPT tends to generate shorter and simpler answers. Without search augmentation, the lengths of SAIL-7B generated sequences are similar to the Vicuna models. This indicates that search augmentation can increase the length of the generated responses.
\begin{table}[t]
\centering
\small
\begin{tabular}{@{}lccc@{}}
\toprule
\textbf{Models}  & \textbf{Avg.} & \textbf{Std.} & \textbf{Diversity} \\ \midrule
GPT-4            & 303.8         & 121.5         & 0.48               \\
ChatGPT          & 135.1         & 63.6          & 0.56               \\
Vicuna-13B       & 204.1         & 82.9          & 0.45               \\
Vicuna-7B        & 196.5         & 90.3          & 0.45               \\
SAIL-7B + Search & 246.2         & 87.7          & 0.44               \\
SAIL-7B          & 206.6         & 86.9          & 0.47               \\ \bottomrule
\end{tabular}
\caption{Statistics about the length and diversity of the generated responses of different language models. Diversity stands for the total number of different words divided by the total length.}
\label{tab:length}
\end{table}


\subsection{Question Answering}
\begin{table*}[h]
\centering
\resizebox{\textwidth}{!}{
\begin{tabular}{@{}lcccccccccccc@{}}
\toprule
\textbf{Model} & \multicolumn{3}{c}{\textbf{LLaMA-7B}} &  \multicolumn{3}{c}{\textbf{Vicuna-7B}} & \multicolumn{3}{c}{\textbf{Vicuna-13B}} & \multicolumn{3}{c}{\textbf{SAIL-7B}} \\
Search   & None             & Wiki & DDG              & None              & Wiki & DDG             & None               & Wiki & DDG             & None         & Wiki & DDG                \\ \midrule
CSQA  & 48.4              & 47.7 & 49.6               & 44.9              & 45.6 & 47.6              & 50.6      & 51.1 &  50.9             & 51.5         & 51.0 & \textbf{51.8}                 \\
OBQA     & 42.2             & 44.4 & 44.6              & 37.2              & 39.4 & 42.6              & 49.0               & 47.2 & 49.4      & 49.2         & 50.2 & \textbf{52.0}                 \\
ARC-C     & 43.0             & 45.2 & 47.3              & 40.5              & 44.5 & 46.3              & \textbf{53.2}               & 51.6 & 51.8     & 47.7         & 48.1 & 48.4                  \\
Avg.     & 44.5             & 45.8 & 47.2              & 40.9              & 43.3 & 45.5              & \textbf{51.0}               & 50.0 & 50.7     & 49.5         & 49.8 & 50.7                  \\ \bottomrule
\end{tabular}
}
\caption{Question answering accuracy (\%) by zero-shot models with simple instructions.}
\label{tab:exp-qa}
\end{table*}
\begin{table*}[]
\centering
\resizebox{\textwidth}{!}{
\begin{tabular}{@{}lcccccccc@{}}
\toprule
\textbf{Model}              & \textbf{Metric} & \textbf{Climate} & \textbf{PubHealth} & \textbf{Fact Avg.} & \textbf{HSD}  & \textbf{SBIC} & \textbf{Fairness Avg.} & \textbf{All Avg.} \\ \midrule
\multirow{2}{*}{Vicuna-7B}  & Acc             & 57.9    & 60.6               & 59.2               & 55.9          & 74.5          & 65.2                   & 62.2              \\
                            & F1              & 38.8             & 56.63              & 47.7               & 68.5          & 84.3          & 76.4                   & 62.04              \\ \midrule
\multirow{2}{*}{Vicuna-13B} & Acc             & 51.4             & 54.4               & 52.9               & 57.7          & 72.3          & 65.0                   & 59.0              \\
                            & F1              & 42.5             & 57.7               & 50.1               & 69.6          & 82.9          & 76.3                   & 63.2              \\ \midrule
\multirow{2}{*}{LLaMA-7B}    & Acc             & 58.8             & 59.9      & 59.3      & 62.3 & 74.8 & 68.6          & 63.9     \\
                            & F1              & 46.6    & 57.5      & 52.0      & 72.3 & 84.4 & 78.4          & 65.2     \\ \midrule
\multirow{2}{*}{SAIL-7B}    & Acc             & \textbf{63.5}             & \textbf{69.2}      & \textbf{66.4}      & \textbf{70.1} & \textbf{76.4} & \textbf{73.2}          & \textbf{69.8}     \\
                            & F1              & \textbf{51.0}    & \textbf{63.6}      & \textbf{57.3}      & \textbf{75.1} & \textbf{83.9} & \textbf{79.5}          & \textbf{68.4}     \\ \bottomrule
\end{tabular}
}
\caption{Instructed zero-shot language checking performance on the UniLC benchmark.}
\label{tab:unilc-0}
\end{table*}

The experiment results of question answering are shown in Table \ref{tab:exp-qa}. CSQA, OBQA, and ARC-Challenge are open-ended, selection-based question-answering tasks. We compare instruction-tuned \texttt{Vicuna-7B}, \texttt{Vicuna-13B}, \texttt{LLaMA-7B-GPT4}, and \texttt{SAIL-7B} models under no-augmentation and search-grounded settings with different sources. All evaluations are zero-shot and instruction guided. Traditionally, a knowledgeable LLM can answer questions and select the most coherent and appropriate answers without external information. In each task, we want to evaluate the performance of different models and knowledge bases. We search Wikipedia (Wiki) with the BM25 retriever, and the web with DuckDuckGO (DDG), feeding the LLMs with the top-3 search results, which could contain unrelated and distracting information.

In general, we found that DuckDuckGo (DDG) leads to better performance for all models on all tasks because it is more flexible, covering a much wider range of information. This suggests the effectiveness of search engines over retrieving a static knowledge base. We found that both \texttt{LLaMA} and \texttt{Vicuna-7B} models can be slightly improved when search results are provided on most tasks. However, the overall performance is limited. The average accuracy of searched-augmented \texttt{LLaMA-7B} and \texttt{Vicuna-7B} is below 50\%.

With \texttt{Vicuna-13B}, which is a roughly two times larger model, we get the best average performance (51.0\%) on the three tasks without grounding information. However, adding search results hurts its accuracy in most experiments. While augmenting the model with DDG search results slightly improves the performance on CSQA and OBQA, the accuracy on ARC-Challenge is decreased by 1.4\%. With BM25-based Wikipedia search results, the accuracy can decrease by as much as 1.8\%. While the \texttt{Vicuna-13B} model achieves strong non-augmented performance, it is challenging to further improve the accuracy by utilizing helpful information in the search results.

In contrast, the \texttt{SAIL-7B} model improves on all tasks when incorporating the search results, and also achieves strong non-augmented performance. Without retrieval results, \texttt{SAIL-7B} significantly outperforms \texttt{LLaMA} and \texttt{Vicuna-7B} on all tasks with a large margin (49.5\% vs 44.5\% and 40.9\% average accuracy). It also performs slightly better than \texttt{Vicuna-13B} on CSQA and OBQA tasks, while Vicuna-13B is still strongest on ARC-C. While search augmentation leads to at most 0.5\% improvement for Vicuna-13B, DDG search results improve \texttt{SAIL-7B} by 2.8\% on OBQA and 1.2\% on average, showing that the \texttt{SAIL-7B} model can steadily utilize the helpful information among the search results. As a result, the search-augmented \texttt{SAIL-7B} model achieves the best performance on both CSQA and OBQA.

\subsection{Fact and Fairness Checking}
\begin{table}[]
\resizebox{\columnwidth}{!}{
\begin{tabular}{@{}llccc@{}}
\toprule
\textbf{Model}              & \textbf{Metric} & \textbf{Climate} & \textbf{PubHealth} & \textbf{Avg.} \\ \midrule
\multirow{4}{*}{Vicuna-7B}  & Acc             & 57.7             & 60.1               & 58.9          \\
& Acc Diff             & -0.2             & -0.5               & -0.3          \\
& F1              & 49.5             & 57.6               & 53.6          \\
                            & F1 Diff              & +10.7             & +1.0               & +5.9          \\ \midrule
\multirow{4}{*}{Vicuna-13B} & Acc             & 53.5             & 50.3               & 51.9          \\
& Acc Diff             & +2.1             & -4.1               & -1.0          \\
                            & F1              & 46.6             & 56.8               & 51.7          \\
                            & F1 Diff             & +4.1             & -0.9               & +1.6          \\\midrule
\multirow{4}{*}{LLaMA-7B}  & Acc             & 55.8             & 62.8               & 59.3          \\
& Acc Diff             & -3.0             & +2.9               & -0.1          \\
                            & F1              & 50.2             & 59.7               & 54.9          \\
                            & F1 Diff              & +3.6             & +2.2               & +2.9          \\ \midrule
\multirow{4}{*}{SAIL-7B}    & Acc             & \textbf{65.8}    & \textbf{70.7}      & \textbf{68.3} \\
& Acc Diff             & +2.3    & +1.5      & +1.9 \\
                            & F1              & \textbf{55.2}    & \textbf{64.5}      & \textbf{59.9} \\
                            & F1 Diff              & +4.2    & +0.9      & +2.5 \\ \bottomrule
\end{tabular}
}
\caption{Search augmented zero-shot language checking performance on the Climate-fever and PubHealth benchmarks.}
\label{tab:fc}
\end{table}
The other task we evaluate model performance on is unified fact and fairness checking \cite{zhang2023interpretable}, a combined benchmark with four sub-tasks including fact-checking \cite{diggelmann2020climate,kotonya-toni-2020-explainable}, hate speech detection \cite{de2018hate}, and stereotype recognition \cite{sap2020social}. We evaluate the zero-shot performance on all four tasks, and the experiment results are shown in Table \ref{tab:unilc-0}. The \texttt{SAIL-7B} model achieves the highest accuracy and F1 scores on all tasks, despite no grounding information being provided for the fact-checking tasks. We also found that the \texttt{Vicuna-7B} and \texttt{13B} models perform similarly on fact and fairness checking.

For the fact-checking tasks, we further evaluate the performance grounding on search results generated by DuckDuckGo. Grounding on an external search engine has both advantages and disadvantages. Many fact checking benchmarks provide task-specific grounding corpora that limit the domain of information retrieval. However, internet misinformation can be very arbitrary and related to the latest facts. A commercial search engine is able to catch a wide range of up-to-date information that a retrieval model with a fixed knowledge base cannot achieve. However, search engines are usually less accurate than dense retrievers, and they might retrieve disputed documents that influence the quality of fact checking. Our experiments show that the search results are not helpful for all models. On Clmate-Fever, augmenting the model with search results decreases the overall accuracy of \texttt{LLaMA} by 3\%. On the PubHealth task, both accuracy and F1 of \texttt{Vicuna-13B} model are decreased by the search results, by 4\% and 1\% respectively. This shows that the search results contain distracting information, which prevents the models to utilize helpful evidence among noises.

However, SAIL is more robust against distracting languages and its fact-checking performance is improved on the same set of search results, as shown in Table \ref{tab:fc}. With search augmentation, the fact-checking accuracy and F1 scores of SAIL are improved on both tasks, as high as 4.2\% on Climate-Fever. The augmented SAIL model also significantly outperforms all baselines, including \texttt{Vicuna-13B} and \texttt{LLaMA-7B} tuned with GPT-4 responses by 9\% accuracy and 5\% F1, showing the effectiveness of search augmented fine-tuning.

\section{Related Work}
\subsection{Capabilities}
\noindent \textbf{Large language models.} Beginning with GPT-3 \cite{NEURIPS2020_1457c0d6}, LLMs have demonstrated strong abilities in knowledge memorization and text-based inference on a wide range of tasks. Well-known LLMs include GPT3, LaMDA \cite{thoppilan2022lamda}, FLAN \cite{wei2021finetuned}, OPT \cite{zhang2022opt}, and LLaMA \cite{touvron2023llama}. Compared to smaller language models, LLMs have several emergent abilities \cite{wei2022emergent}, including zero-shot multi-task solving, and few-shot in-context learning with chain-of-thought reasoning \cite{wei2022chain,wang2022self}.\\

\noindent \textbf{Instruction following.} Pretrained LLMs can generate texts following certain formats and rules by seeing a few examples in their prompts. To make LLMs more scalable and improve zero-shot performance, \citet{ouyang2022training} proposed training GPT3 with instruction-response corpora. As a result, InstructGPT, ChatGPT, and GPT4 can handle a wide range of tasks without seeing any examples. Recent research has also found that both GPT-generated instructions and instruct-following outputs \cite{peng2023instruction} can improve the instruction-following ability of LLMs. \cite{wang2022self} proposed a semi-supervised method to generate diverse instructions based on a seed instruction base on NLP tasks \cite{naturalinstructions,supernaturalinstructions}. A more recent study shows that GPT-4 \cite{openai2023gpt4} can generate high-quality instruction-following language. Recent efforts on open-sourcing instruction-following LLMs include Alpaca \cite{alpaca} and Vicuna \cite{vicuna2023}.\\

\noindent \textbf{Retrieval-augmented language models.} 
Prior to our work, several initiatives explored retrieval-augmented language models (RALMs). The pioneering approaches -- REALM~\cite{guu2020retrieval} and RAG~\cite{lewis2020retrieval} -- sought to train language models with retrievers in an end-to-end manner. RETRO~\cite{borgeaud2022improving} introduced the idea of training an LM on top of a frozen retriever. Atlas~\cite{izacard2022few} further explored dedicated loss functions for the end-to-end training of the retriever and the LM, achieving superior performance on several few-shot learning tasks. Recently, RePlug~\cite{shi2023replug} and In-context RALM~\cite{ram2023context} instead explore an opposite direction: use a frozen black-box LM while fine-tuning the retrieval modules. RePlug shows its advantages of leveraging large LMs like Codex~\cite{chen2021evaluating} and GPT-3~\cite{brown2020language}, outperforming Altas on few-shot question-answering tasks.

Despite the success of RALMs, most of these models have limitations, including 1) constraining the search space to a closed corpus like Wikipedia 2) lacking explicit mechanisms for disregarding distracting search results, and 3) applying a few-shot in-context learning setting without considering instruction fine-tuning during RALM training. Consequently, their applications remain relatively narrow, primarily focusing on tasks such as question-answering and language modeling. SAIL addresses these limitations by 1) employing real-world search engines, 2) introducing a search result denoising process capable of filtering out distracting information, and 3) incorporating instruction fine-tuning. Consequently, SAIL demonstrates its superiority in broader applications, including instruction following for chatbots, fact and fairness checking, all of which benefit from access to up-to-date information retrieved from real-world search engines.

\subsection{Trustworthiness}

\noindent \textbf{Self-improving.} Recent studies have found that both pretrained and instruction fine-tuned LLMs can improve themselves with appropriate prompting strategies. Compared to directly generating the answers, the step-by-step, chain-of-thought \cite{wei2022chain} generation strategy significantly improves the reasoning accuracy. Furthermore, self-consistent predictions are usually more trustworthy \cite{wang2022self}.  \citet{huang2022large} showed that self-consistent predictions generated by LLMs can be used as in-context examples that significantly improve task and domain adaptation. After instruction fine-tuning, language models can generate suggestions to improve their own outputs with self-reflection and self-refinement prompting strategies \cite{shinn2023reflexion,madaan2023self}. \\

\noindent \textbf{Fact and fairness checking.} Aside from an ability to generate correct responses, we believe that LLMs should take the responsibility of checking undesirable and harmful language generated by both machines and humans. \citet{manakul2023selfcheckgpt} found that the GPT-3 model can identify its own hallucinations, and \citet{zhang2023interpretable} proposed a unified fact and fairness checking framework for both human and machine-generated language.

\section{Conclusion}
In this work, we found that disputed and distracting search results can significantly mislead the predictions of large language models. Several transparency-sensitive tasks, including open-domain question answering and language checking can be negatively influenced by this phenomenon. To solve this problem, we propose a search-augmented instruction-following large language model with 7B parameters. We construct the first search-augmented instruction-tuning corpus consisting of human-generated instructions, GPT-4 generated responses, and search results generated by a BM25 retriever based on Wikipedia and a commercial search engine. We then fine-tuned the \texttt{LLaMA-7B} language model with the constructed training corpus on academic computational resources. Experiments on instruction-following, question answering, and fact/fairness checking show that the search-augmented language model can distill trustworthy and helpful information from all search results and generate high-quality responses, improving both the performance and transparency of instruction-following large language models.

\section*{Acknowledgement}
This research was supported by the Center for Perceptual and Interactive Intelligence (CPII) Ltd under the Innovation and Technology Commission’s InnoHK Scheme.

\section*{Limitations}
While the model we propose achieves high performance with efficient model settings, the major limitation of the model is that it does not explain why a search result is trustworthy or informative or not. In future work, we will fine-tune larger models and enable the models to recognize trustworthy search results with explanations.

\bibliography{anthology,custom}
\bibliographystyle{acl_natbib}




\end{document}